\newif\ifarxivstyle
\arxivstyletrue 

\ifarxivstyle
    \documentclass{article}
    \usepackage{arxiv}
\else
    \documentclass[runningheads]{llncs}
\fi
    
\usepackage[T1]{fontenc}
\usepackage{newtxtext}
\usepackage{amsmath, amssymb, bm, dsfont}
\usepackage[varvw]{newtxmath}
\usepackage[mathscr]{euscript}

\usepackage{graphicx, tikz, tikzscale}
\usepackage{adjustbox, rotating, placeins}
\usepackage[caption=false]{subfig}
\usetikzlibrary{
    arrows, arrows.meta, positioning, shapes,
    backgrounds, patterns, calc,
    decorations.pathreplacing, decorations.pathmorphing
}

\usepackage{array, tabularx, booktabs, colortbl}
\usepackage{algorithm, algpseudocode}
\newcolumntype{Y}{>{\centering\arraybackslash}X}

\usepackage{microtype}
\usepackage{xcolor, contour, soul}
\usepackage[normalem]{ulem}
\usepackage{verbatim, mdframed, tcolorbox}

\usepackage[acronym]{glossaries}
\newacronym{dt}{DT}{Decision Tree}
\newacronym{ea}{EA}{Evolutionary Algorithm}
\newacronym{ec}{EC}{Evolutionary Computation}
\newacronym{gp}{GP}{Genetic Programming}
\newacronym{llm}{LLM}{Large Language Model}
\newacronym{lr}{LR}{Linear Regression}
\newacronym{ml}{ML}{Machine Learning}
\newacronym{mae}{MAE}{Mean Absolute Error}
\newacronym{mse}{MSE}{Mean Squared Error}
\newacronym{nn}{NN}{Neural Network}
\newacronym{qr}{QR}{Quadratic Regression}
\newacronym{rbf}{RBF}{Radial Basis Function}
\newacronym{svm}{SVM}{Support Vector Machines}
\newacronym{sr}{SR}{Symbolic Regression}
\newacronym{xgboost}{XGBoost}{eXtreme Gradient Boosting}
\newacronym{aml}{AML}{Adversarial Machine Learning}

\usepackage{hyperref,orcidlink}

\newif\ifshowcomments
\showcommentsfalse

\newcommand{\authorcomment}[3]{%
    \ifshowcomments
    \par\noindent
    \begin{tcolorbox}[
        colback=#1!15,
        colframe=#1!70!black,
        boxrule=0.6pt,
        arc=2pt,
        left=3pt,
        right=3pt,
        top=2pt,
        bottom=2pt,
        title={\footnotesize\textbf{#2's comment:}},
        fonttitle=\sffamily
    ]
        \footnotesize\sffamily #3
    \end{tcolorbox}%
    \fi
}

\newcommand{\ak}[1]{\authorcomment{green}{Anna}{#1}}
\newcommand{\alberto}[1]{\authorcomment{cyan}{Alberto}{#1}}

\newif\iffinalversion
\finalversiontrue

\begin{document}
\title{Evolutionary Generation of \\Adversarial Regression Datasets}
\title{Beyond Adversarial Samples:\\Evolving Adversarial Datasets for Tabular Regression} 
\title{Evolving Adversarial Datasets for Tabular Regression}
\title{Evolutionary Adversarial Dataset Generation\\Uncovers Issues in XGBoost}

\ifarxivstyle
    \author{ \href{https://orcid.org/0000-0002-3136-6502}{\includegraphics[scale=0.06]{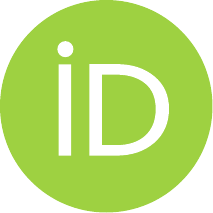}\hspace{1mm}Omar~Lopez-Rincon} \\
    Applied Mathematics and Computer Science, Paris-Saclay (MIA-PS)\\
	Université Paris-Saclay, INRAE\\
	22 place de l'Agronomie, 91120, Palaiseau, France\\
	\texttt{omar.lopez-rincon@inrae.fr} \\
	\And
	\href{https://orcid.org/0000-0002-4138-7024}{\includegraphics[scale=0.06]{orcid.pdf}\hspace{1mm}Anna~V.~Kononova} \\
    LIACS\\
	Leiden University\\
	Rapenburg 70, 2311 EZ Leiden, Netherlands\\
	\texttt{a.kononova@liacs.leidenuniv.nl} \\  
	\And
	\href{https://orcid.org/0000-0001-5895-4809}{\includegraphics[scale=0.06]{orcid.pdf}\hspace{1mm}Alberto~Tonda} \\
    Applied Mathematics and Computer Science Paris-Saclay (MIA-PS)\\
	Université Paris-Saclay, INRAE\\
	22 place de l'Agronomie, 91120, Palaiseau, France\\
    Complex Systems Institute of Paris\\
    CNRS\\
    113 rue Nationale, 75013, Paris, France\\
	\texttt{alberto.tonda@inrae.fr} \\
    }
    \hypersetup{
    pdftitle={Evolutionary Adversarial Dataset Generation Uncovers Issues in XGBoost},
    pdfsubject={cs.NE,cs.LG}, 
    pdfauthor={Omar~Lopez-Rincon, Anna~V.~Kononova, Alberto~Tonda},
    pdfkeywords={Adversarial machine learning, genetic programming, machine learning, support vector machines, symbolic regression, XGBoost},
    }
\else
    \iffinalversion
        \author{
        Omar Lopez-Rincon\inst{1}\orcidlink{0000-0002-3136-6502}  \and
        Anna V. Kononova\inst{2}\orcidlink{0000-0002-4138-7024} \and
        Alberto Tonda\inst{1,3}\orcidlink{0000-0001-5895-4809}
        }
        
        \authorrunning{O. Lopez, A. Tonda and A.V. Kononova}
        
        \institute{
        Université Paris-Saclay, INRAE, Applied Mathematics and Computer Science\\Paris-Saclay (UMR 518 MIA-PS), France\\
        \email{\{omar.lopez-rincon,alberto.tonda\}@inrae.fr} 
        \and
        LIACS, Leiden University, The Netherlands\\
        \email{a.kononova@liacs.leidenuniv.nl} 
        \and
        CNRS, Complex Systems Institute of Paris Ile-de-France (UAR 3611 ISC-PIF), France
        }
    \fi
\fi

\maketitle 

\begin{abstract}
Machine Learning (ML) algorithms have been widely successful, but the complexity of trained models makes their behavior hard to analyze and predict, and their weaknesses difficult to uncover. Adversarial ML addresses this challenge by generating single samples that a victim algorithm mispredicts, while keeping such adversarial samples as close as possible to the original training data. While useful for detecting failure points, this approach cannot provide information on general algorithmic behavior. We propose a novel approach to adversarial dataset generation for tabular regression, aiming to minimize the performance of a victim algorithm while keeping the explicit function encoded in the dataset learnable by a reference algorithm. Such a function can then be studied to uncover hidden weaknesses of the algorithm. A Genetic Programming framework generates candidate functions, which are then sampled. Victim and reference algorithms are applied on the resulting datasets in a $k$-fold cross-validation, with the objective of lowering the \emph{best} performance of the victim over the folds while keeping the reference's \emph{worst} performance as high as possible. A first experiment on a trivial benchmark, with Linear Regression as the victim and Quadratic Regression as the reference, returns non-linear expressions, as expected. A second experiment with two state-of-the-art algorithms, XGBoost as the victim and Support Vector Machines as the reference, identifies a function on which XGBoost’s performance suffers considerably. Analyzing the function, we uncover undocumented undesirable behavior in XGBoost’s library, used by practitioners worldwide, showing the practical relevance of the proposed approach.

\end{abstract}

\section{Introduction}

Given a dataset with $S$ samples $\mathcal{D}_f=\{(\mathbf{x}_i, y_i)\}_{i=1}^S$, generated by sampling an unknown function $f$, a supervised \gls{ml} algorithm $a$ learns a predictive function $\hat{f}_a$ so that $\hat{f}_a(\mathbf{x}_i) \approx y_i, \forall i$. \gls{ml} algorithms have shown impressive effectiveness in obtaining predictive models starting from data in real-world problems. However, complexity of these algorithms, both in structure and number of parameters, typically makes them effectively \emph{black-box}; humans find it is difficult to interpret their behavior and even understand their weaknesses. This is not only a theoretical problem, but a practical issue of great importance: For example, despite several works on the subject~\cite{barbiero2020modeling,palumbo2023algorithm,prudencio2011selecting}, at the time of writing, it is still next to impossible to answer the question ``\textit{which \gls{ml} algorithm, and which hyperparameter configuration, are best suited to a given dataset?}''. In practice, often users approach model selection with a suite of algorithms, performing hyperparameter tuning for the most promising ones; such a procedure requires a considerable amount of computational time and might still fail to locate the optimal configuration. 

Other attempts of studying the behavior of \gls{ml} algorithms, in particular tabular regressors, focused on analyzing their behavior during training in \emph{semantic space}~\cite{moraglio2012geometric}. A point $\mathbf{\hat{y}_a}$ in semantic space corresponds to the predictions of a specific trained regressor $\hat{f}_a$ for all training samples $S$, $\mathbf{\hat{y}_a}=(\hat{f}_a(\mathbf{x}_1), \dots,\hat{f}_a(\mathbf{x}_S)), \mathbf{\hat{y}}_a \in \mathbb{R}^S$. By observing a sequence of points in semantic space, for example during subsequent training iterations of a \gls{nn} or \gls{xgboost}, it is possible to find commonalities and differences. In practice, since the dimensionality $S$ of the semantic space can be considerably high, trajectories are projected in 2D spaces, what is commonly known as \emph{behavioral} space~\cite{santos2026visualizing,santos2027semanticviz}.


Furthermore, while it is known that \gls{ml} algorithms suffer from overfitting, it is extremely hard to predict which samples will be the most problematic. The sub-field commonly known as \emph{adversarial} \gls{ml} tackled this problem by attempting to generate samples minimally different from what was observed in training~\cite{szegedy2014intriguing}, mostly targeting Deep \glspl{nn} and using other \glspl{nn} to create the samples~\cite{goodfellow2014generative}. \glspl{ea} have also been employed for adversarial \gls{ml}~\cite{su2019one,Zarzecki2026}, again with the objective of finding single adversarial samples which would induce errors in a trained model. Interestingly, in an unrelated domain, which could now be also retrospectively labeled \emph{adversarial}, \gls{sr} has been used to generate functions designed to be difficult to optimize for a target algorithm~\cite{Langdon2007}, in order to study the weaknesses of different optimization algorithms.

Taking inspiration from these earlier works, we propose an evolutionary approach for the generation of entire tabular adversarial regression datasets, based on generated closed-form functions, which best differentiate the performance of two regressors. After choosing a \gls{ml} regressor as \textit{victim} $v$, we evolve \gls{sr} functions $f$ that are then sampled to obtain a dataset $\mathcal{D}_f=\{(\mathbf{x}_i,y_i)\}_{i=1}^S$ with $S$ samples. The objective is to minimize the performance of $v$, all the while keeping the performance of another \textit{reference} regressor $r$ as high as possible, to ensure that the functions $f$ are actually learnable with the given sampling. The resulting functions can be then studied to analyze the behavior of $v$ and potentially uncover previously unknown weaknesses. Furthermore, $f$ can potentially be sampled for different values of $\mathbf{x}_i$, to obtain more instances of the same problem. 

A first set of experiments on a trivial benchmark, with \gls{lr} as the victim and \gls{qr} as the reference, shows that the proposed approach produces expressions which can be approximated by quadratic polynomials, as expected. The reverse (using \gls{qr} as the victim and \gls{lr} as the reference) predictably does not generate useful results. A second set of experiments, this time using two state-of-the-art regressors, \gls{xgboost} as the victim and \gls{svm} as the reference, finds a simple expression which surprisingly leads \gls{xgboost} to an absymal performance, while being easy to learn for \gls{svm}. Investigating the code of \gls{xgboost}'s library, which is currently the most commonly used in the world, reveals an undocumented \textit{bug} which generates the undesired behavior, thus providing a first real-world validation of the proposed approach. This approach paves the way for a systematic investigation of the relative strengths and weaknesses of ML algorithms.


The rest of the paper is organized as follows. Section~\ref{sec:background} introduces the minimal background elements necessary to understand the scope of this work. The proposed approach is presented in Section~\ref{sec:proposed-approach}, and the experimental evaluation is described in Section~\ref{sec:experimental-evaluation}. Section~\ref{sec:limitations} discusses the limitations of the framework, and Section~\ref{sec:conclusions} concludes the paper and outlines future directions for this research line.

\alberto{I like to reset acronym definitions at the end of the introduction, let me know if you agree with this.}
\glsresetall
\ak{If you really insist.. I actually find it quite annoying to have too many re-introduced abbreviations, as it makes the text look auto-generated.}

\section{Background} \label{sec:background}
\subsection{Machine learning for tabular regression}
Tabular regression is the task of predicting a continuous target $y$ from data organized in samples, where each sample is an observation of features $\mathbf{x_i}=(x_{i,1},\dots,x_{i,d}) \in \mathbb{R}^d$; a regression algorithm $a$ trained on a dataset $\mathcal{D}=\{(\mathbf{x}_i, y_i)\}_{i=1}^S$ should be able to learn a function $\hat{f}_a$ so that $y_i \approx \hat{f}_a(\mathbf{x}_i), \forall i$. Examples include predicting house prices from property attributes, a patient’s risk score from clinical measurements, or energy demand from weather and calendar variables. Data in finance, healthcare, science, and industry is typically stored in tables, so tabular regression and classification are often claimed to be the most widely deployed forms of \gls{ml} in practice~\cite{shwartz-ziv2022tabular}.

Given a \gls{ml} algorithm $a$, its \emph{hypothesis space} $\mathcal{H}_a$ represents the set of all functions $a$ can possibly choose from when fitting data. For example, \gls{lr}’s hypothesis space is all linear functions of the inputs in the form $\hat{f}(\mathbf{x})=\beta_0 + \beta_1 \cdot x_1 + \beta_2 \cdot x_2 + \dots + \beta_d \cdot x_d$, fully defined by $\beta_i,i\in\{0,...,d\}$. The training process can then be seen as a search through $\mathcal{H}_a$ for the hypothesis that best fits the data, according to a given objective function. The hypothesis space is also connected to the concept of \emph{capacity}, a measure of how large or expressive $\mathcal{H}_a$ is or, in other words, the variety of functions the model can represent. \gls{lr} has low capacity; a deep \gls{nn} with millions of parameters will have an extremely high capacity. In some cases, it is trivial to assess relative capacity: For example, \gls{qr}, which is essentially a \gls{lr} with extra features computed from the original data, has higher capacity than \gls{lr}, as $\mathcal{H}_{LR} \subset \mathcal{H}_{QR}$. If algorithm $a$'s $\mathcal{H}_a$ is too small (low capacity), it may not contain anything close to the true underlying function, leading to \emph{underfitting}: Poor performance on both training and test data. If $\mathcal{H}_a$ is too large (high capacity), the model can find hypotheses that fit the training data perfectly, including its noise, going into \emph{overfitting}: Low training error but poor generalization to unseen data, such as a test set. In practical cases, estimating capacity can be extremely difficult, with theoretical frameworks such as the Vapnik-Chervonenkis dimension~\cite{vapnik2015uniform} only able to define loose upper bounds for state-of-the-art models, such as deep \glspl{nn} and ensembles of trees.

Tree-based models are currently considered among the most competitive algorithms for tabular regression~\cite{grinsztajn2022why,shwartz-ziv2022tabular}, even though recent Transformer-based approaches (such as TabPFN) have reported competitive or superior performance~\cite{hollmann2023tabpfn}. Tree-based ensembles use \glspl{dt}~\cite{breiman1984classification} as base predictors. A \gls{dt} addresses regression by recursively partitioning the feature space into axis-aligned regions through a sequence of feature-threshold tests. Each terminal leaf is associated with a prediction value. Splits are selected to minimize a loss or impurity criterion, but unconstrained trees can easily overfit unless regularized through depth limits, minimum leaf size or pruning. To improve performance, modern approaches rely on ensembles of \glspl{dt}. \gls{xgboost}~\cite{chen2016xgboost}, commonly considered one of the best regression algorithms by practitioners, builds trees sequentially, with each new tree trained to correct the residual errors of the current ensemble by optimizing a differentiable loss function. It combines regularization with efficient optimization strategies to improve predictive accuracy while controlling overfitting.

Another algorithm considered among the state of the art for regression is \gls{svm}~\cite{cortes1995supportvector}. When used for regression, \gls{svm} fits a function that keeps as many training points as possible inside a tube of radius $\epsilon$ around the prediction. Errors smaller than $\epsilon$ are ignored entirely; only points that fall on or outside the tube become support vectors and shape the solution, which makes the model sparse and fairly robust to small noise. The optimization balances two goals: keeping the function as flat as possible (small weights, for good generalization) and penalizing points outside the tube, with the regularization parameter $C$ controlling that trade-off. Through the kernel trick, using kernels such as \gls{rbf} or polynomial, \gls{svm} can capture non-linear relationships by implicitly working in a higher-dimensional feature space.

\subsection{Adversarial machine learning}
\gls{aml} emerged in the early 2000s from research on classification security. Dalvi et al.~\cite{dalvi2004adversarial} modeled classification as a game between a classifier and an adversary modifying inputs to evade detection, motivated by spam filtering. Subsequent work investigated evasion and poisoning attacks against conventional ML algorithms, including support vector machines. Biggio et al.~\cite{biggio2013evasion} demonstrated gradient-based evasion attacks, while Biggio and Roli~\cite{biggio2018wild} review the early developments of the field.

\gls{aml} gained widespread attention when Szegedy et al.~\cite{szegedy2014intriguing} showed that small, often visually subtle perturbations could cause deep neural networks to misclassify images, sometimes transferring across models. Goodfellow et al.~\cite{goodfellow2015explaining} attributed this vulnerability partly to the linear behavior of high-dimensional models and introduced the Fast Gradient Sign Method. Subsequent work developed stronger attacks~\cite{carlini2017towards} and defenses such as \textit{adversarial training}, formulated as a min--max optimization problem~\cite{madry2018towards}.

When an attack cannot rely on access to a model's internal structure or gradients, \gls{ec} can be used to find \textit{adversarial inputs} using only the model's predictions. Nguyen et al.~\cite{nguyen2015deep} used \glspl{ea} and gradient ascent to generate unrecognizable images classified with high confidence by neural networks. Su et al.~\cite{su2019one} used Differential Evolution for one-pixel attacks, while Alzantot et al.~\cite{alzantot2019genattack} developed query-efficient attacks using Genetic Algorithms. Beyond images, Xu et al.~\cite{xu2016automatically} used genetic programming to evolve malicious PDF documents that evaded classifiers while preserving malicious functionality. More recently, Panfilov et al.~\cite{Panfilov2026} used LLM agents to automate adversarial attack design.

Similar approaches exist in \gls{ml} benchmarking. Wilde et al.~\cite{Wilde2019} introduced Evolutionary Dataset Optimization to generate datasets favoring particular learning algorithms. Shand et al.~\cite{Shand2022} developed HAWKS to evolve clustering benchmarks, including datasets maximizing performance differences between pairs of algorithms. Orzechowski and Moore~\cite{Orzechowski2022} proposed DIGEN, which evolves classification problems to expose differences between classifiers.

A related direction in \gls{ec} is the generation of \textit{challenging problem instances} rather than adversarial inputs. Langdon and Poli~\cite{Langdon2007} evolved optimization problems to maximize performance differences between algorithms. Bossek and Wagner~\cite{Bossek2021} extended this approach to more than two algorithms. Hajari and Gallagher~\cite{Hajari2024} generated challenging clustering-based optimization instances through data perturbations, while Hart et al.~\cite{Hart2024} evolved bin-packing instances that misled deep-learning-based algorithm selectors. Ono et al.~\cite{Ono2026} recently proposed using LLMs within an evolutionary framework to generate continuous optimization benchmarks favoring one algorithm over another. Zarzecki and Arabas~\cite{Zarzecki2026} explored a complementary direction, proposing black-box adversarial attack problems as challenging benchmarks for evaluating global optimization algorithms. These studies also relate to a broader research agenda concerned with understanding and \textit{explaining the interplay} between problem characteristics and algorithm behavior~\cite{xro2026}.


\subsection{Genetic Programming and Symbolic Regression}
\gls{gp}, originally proposed in~\cite{koza1994genetic}, is an \gls{ea} performing optimization over tree structures, which, depending on the choice of operator and terminal sets, can represent a wide range of structures, from computer programs to mathematical equations. One of the most successful applications of \gls{gp} is \gls{sr}, a \gls{ml} algorithm typically used for tabular regression problems, able to find closed-form equations which fit a given training dataset. Modern approaches to \gls{sr}~\cite{cranmer2023interpretable,Schmidt2009} often deliver a complexity/error Pareto front of non-dominated solutions, leaving the final choice of expression to the user~\cite{smits2005paretofront}. 

\section{Proposed approach}\label{sec:proposed-approach}

\begin{figure}[tb]
    \centering
    \includegraphics[width=0.999\textwidth, trim=0cm 2cm 0cm 0cm]{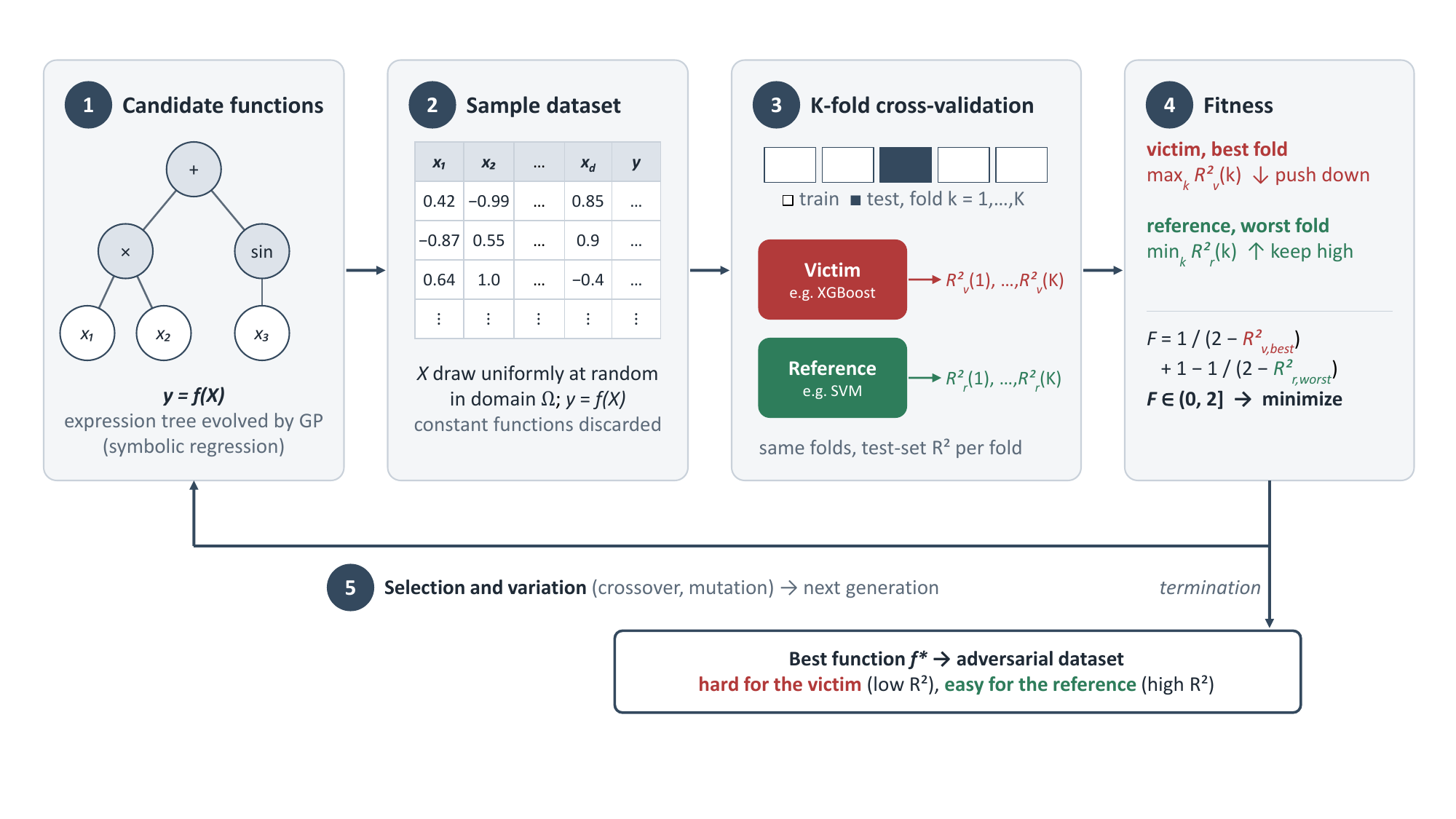}
    \caption{Overview of the proposed approach. Candidate functions $f$ (1), represented as \gls{gp} trees, are sampled (2). A victim \gls{ml} algorithm $v$ and a reference $r$ are trained in a $k$-fold cross-validation on the resulting dataset (3). The fitness value of each candidate function is linked to the best performance of $v$ and the worst performance of $r$ (4). Candidate solutions are evolved (5) with the aim of obtaining a function difficult to learn for $v$, but that $r$ can still approximate efficiently.}
    \label{fig:overview}
\end{figure}

Taking inspiration from~\cite{Langdon2007} and the literature on adversarial \gls{ml}, we propose a novel approach designed to generate entire adversarial tabular \emph{datasets}, instead of single samples. Given a victim regressor $v$, we optimize a symbolic function $f$ to be as difficult as possible for $v$ to learn, while still being predictable by a different reference regressor $r$. Once a function $f^*$ that maximizes the performance difference between the two regressors in favor of $r$ is found, its symbolic expression can be analyzed to uncover possible weaknesses of the victim. Furthermore, since the method produces a symbolic expression and not just a single tabular dataset, $f^*$ can be \textit{sampled again to obtain additional datasets} representing the same underlying regression problem. An overview of the approach is presented in Figure~\ref{fig:overview}.


\subsection{Generating and evolving candidate functions}\label{sect:generating}
Candidate functions are encoded with classical \gls{gp} trees for \gls{sr}~\cite{koza1994genetic}. Let $\mathcal{F}$ be the set of all expressions generated by the \gls{gp} grammar, with \textit{primitives} such as $\{+,-,\times,\div,\sin,\cos,\exp,\dots\}$, \textit{features} $x_1,\dots,x_d$ and \textit{constants}. Each individual will then encode an expression $f \in \mathcal{F}$. The approach proposed here can rely upon \textit{any} \gls{gp} algorithm, as the main novelty lies in the fitness function: During the experimental evaluation, we employ a state-of-the-art \gls{sr} library~\cite{cranmer2023interpretable}, but in principle the engine can be replaced by any other framework able to generate symbolic expressions and efficiently explore the search space $\mathcal{F}$ guided by an objective function.

\subsection{Fitness function guiding the evolution of \gls{sr} expressions}
Let $f:\Omega\subseteq\mathbb{R}^d\to\mathbb{R}$ be a candidate \gls{sr} expression, defined on $\Omega = [-1,1]^d$. This function is sampled for $S$ inputs $\mathbf{x}_i \sim \mathcal{U}(\Omega)$ i.i.d., to build dataset $\mathcal{D}_f=\{(\mathbf{x}_i, y_i)\}_{i=1}^{S}$ with $y_i=f(\mathbf{x}_i)$. Dataset $\mathcal{D}_f$ can then be used to assess the capability of \gls{ml} regression algorithm $a$ to learn an approximation $\hat{f}_a$ of $f$ from the samples.

The performance of a regression algorithm can be evaluated resorting to metrics such as \gls{mse} or \gls{mae}. Here, we decide to employ $R^2 \in (-\infty,1.0]$ for its desirable numerical bounds and easier interpretability: Values close to $1.0$ correspond to \textit{perfect} predictions, values close to $0.0$ or negative are \textit{unsatisfactory}. 

When evaluating the performance of a regression algorithm, employing the same dataset used to train it would greatly overestimate its performance, due to overfitting issues. A classical solution is to randomly split the dataset into a training and a test set, and evaluate performance on test; however, in practice, it is much more reliable to split the data into folds and evaluate the mean performance over $K$ test sets, a procedure known as $k$-fold cross-validation~\cite{geisser1975predictive}. $\mathcal{D}_f$ is thus partitioned into $K$ folds; let $T_k\subset\{1,\dots,S\}$ be the indices of the test samples of fold $k$, and $\hat{f}_a^{(k)}$ the model
fitted by algorithm $a\in\{v,r\}$ on $\mathcal{D}_f\setminus T_k$. The coefficient of determination $R^2_a(k)$ of model $\hat{f}_a^{(k)}$ on fold $k$ will thus be:
\begin{equation}
\begin{split}
  R^2_a(k) &= \begin{cases}1 - \frac{\sum\limits_{i\in T_k}\bigl(y_i-\hat{f}_a^{(k)}(\mathbf{x}_i)\bigr)^2} 
                      {\sum\limits_{i\in T_k}\bigl(y_i-\bar{y}_{k}\bigr)^2} = 1 - \frac{\sum\limits_{i\in T_k}\bigl(f(\mathbf{x}_i)-\hat{f}_a^{(k)}(\mathbf{x}_i)\bigr)^2} 
                      {\sum\limits_{i\in T_k}\bigl(f(\mathbf{x}_i)-\bar{y}_{k}\bigr)^2} & \text{if } \sum\limits_{i\in T_k}\bigl(y_i-\bar{y}_{k}\bigr)^2 \ne 0\\
                      0 & \text{otherwise}
  \end{cases}\\
  & \text{with } \bar{y}_{k}=\frac{1}{|T_k|}\sum_{i\in T_k} f(\mathbf{x}_i)
  \label{eq:r2}
\end{split}
\end{equation}

For our approach, the mean performance $\overline{R^2} = \sum^K_{k=1}{\frac{R^2(k)}{K}}$ over the $K$ folds could be too sensitive to, for example, a single fold where the victim-reference performance vastly differed. To reduce sensitivity to random splits over the folds, we consider the \emph{best} single-fold performance obtained by the victim and the \emph{worst} single-fold performance obtained by the reference. The final equation for the fitness function $F$ is thus:

\begin{equation}
\begin{split}
F(f) &= 1 + \underbrace{\frac{1}{2-\underset{k}{\max} R^2_v(k)}}_{\text{victim's best performance}}
       - \underbrace{\frac{1}{2-\underset{k}{\min} R^2_r(k)}}_{\text{reference's worst performance}} =\\
    &= 1 + \frac{1}{2-\underset{k}{\max} \left( 1 - \frac{\sum\limits_{i\in T_k}\bigl(f(\mathbf{x}_i)-\hat{f}_v^{(k)}(\mathbf{x}_i)\bigr)^2} 
                      {\sum\limits_{i\in T_k}\bigl(f(\mathbf{x}_i)-\bar{y}_{k}\bigr)^2}\right)} - \frac{1}{2-\underset{k}{\min} \left( 1 - \frac{\sum\limits_{i\in T_k}\bigl(f(\mathbf{x}_i)-\hat{f}_r^{(k)}(\mathbf{x}_i)\bigr)^2} 
                      {\sum\limits_{i\in T_k}\bigl(f(\mathbf{x}_i)-\bar{y}_{k}\bigr)^2}\right)}\\
    & \text{if } \sum_{i\in T_k}\bigl(f(\mathbf{x}_i)-\bar{y}_{k}\bigr)^2 \ne 0
\label{eq:fitness-function}
\end{split}
\end{equation}

And the optimization problem can then be defined as:
\begin{equation}
f^* = \underset{f\in\mathcal{F}}{\arg\min}\; F(f).
\end{equation}

The numerical constants in Eq.~\ref{eq:fitness-function} have values selected so that $F \in [0.0,2.0]$ and for $\underset{k}{\max} R^2_v(k) \rightarrow -\infty$, and $\underset{k}{\min} R^2_r(k) \rightarrow 1.0$, $F \rightarrow 0.0$. Looking back at Eq.~\ref{eq:r2}, one might notice how a constant function $f_c(\mathbf{x})=c$ would lead the denominator $\sum_{i\in T_k}\bigl(f_c(\mathbf{x}_i)-\bar{y}_{k}\bigr)^2=0$, and result in $R^2_v(k)=R^2_r(k)=0, \forall k$, with $F(f_c)=1.0$; however, these functions are trivial to learn for all modern regression algorithms, therefore constant candidate functions will be \textit{discarded before evaluation}.

\section{Experimental evaluation}
\label{sec:experimental-evaluation}
The proposed approach has been implemented in \texttt{Julia}, and all the necessary code to reproduce the experiments is available in a public GitHub repository\footnote{Anonymized for double-blind submission, \url{https://anonymous.4open.science/r/adversarial-datasets-ml-5383/}
}. The \gls{gp} engine employed is the state-of-the-art library \texttt{SymbolicRegression.jl}~\cite{cranmer2023interpretable}. For all experiments, the \gls{sr} algorithm is set with the following hyperparameters: operator set $\{+,-,*,/,\sin,\cos,\log,\exp\}$, population size 50 (for each of the 31 subpopulations), maximum number of generations 100, maximum equation complexity (number of nodes in the \gls{gp} tree) 30. The regressors used are implemented by \texttt{XGBoost.jl} and \texttt{LIBSVM.jl}~\cite{chang2011libsvm}, and for each regressor we employ its library's default hyperparameter values: \gls{xgboost} uses $n=100$ predictors and \gls{svm} employs an \gls{rbf} kernel, $C=1.0$, $\epsilon=0.1$. Unless stated otherwise, the number of features is set to $d=3$, so each candidate function $f:\Omega\subseteq\mathbb{R}^3\to\mathbb{R}$ is defined over $\Omega=[-1,1]^3$; the number of samples i.i.d. extracted from $\Omega$ is set to $S=200$. The resulting dataset is generated once at the beginning of each experiment and randomly split into $k=5$ folds, using a fixed random seed. All experiments have been run on a machine with an Intel\textregistered Xeon\textregistered~CPU E5-2630 v3 @ 2.40GHz and 128GB of RAM.

Each experimental configuration is run 3 times, and the resulting Pareto fronts are aggregated, with the final results displaying only the non-dominated candidate functions.

\subsection{Linear vs Quadratic regression}
A simple test for the proposed approach is to verify whether it can discover a result we expect. Estimating the capacity of \gls{ml} algorithms is generally quite complex~\cite{vapnik2015uniform}, but in the case of polynomial regression it is straightforward: For example, \gls{lr} will have a lower capacity than \gls{qr}, as $\mathcal{H}_{LR} \subset \mathcal{H}_{QR}$. In other words, any function representable by \gls{lr} can also be represented by \gls{qr} by setting all second-order coefficients to zero, while the reverse is not true.

{\renewcommand{\arraystretch}{1.1}\setlength{\tabcolsep}{6pt}\begin{table}[!htb]
    \centering
    \caption{Candidate equations from the Pareto front obtained with \gls{lr} as the victim and \gls{qr} as the reference.}
    \label{tab:pareto-lr-qr}
    \resizebox{\textwidth}{!}{%
    \begin{tabular}{l|c|c|c|c|c|c}
        Candidate function $f(\mathbf{x})$ & Complexity & $F$ &  $R^2_{LR}$ (best) & $R^2_{QR}$ (worst) & $\overline{R^2}_{LR}$ &  $\overline{R^2}_{QR}$ \\
        \hline \hline
        $x_{1}$ & 1 & 1.0 & 1.0 & 1.0 & 1.0 & 1.0 \\
        \hline
        $\cos( x_{1} )$ & 2 & 0.4950 & -0.0227 & 0.9994 & -0.0543 & 0.9995 \\ \hline
        \begin{tabular}[c]{@{}l@{}}$( x_{2} \cdot x_{2} ) \cdot ( ( 0.46526137 / ( ( x_{3} \cdot x_{2} ) \cdot x_{2} ) ) \cdot ( ( x_{3}$ \\ \quad $\cdot x_{2} ) \cdot x_{2} ) )$\end{tabular} & 17 & 0.4939 & -0.0249 & 1.0 & -0.0548 & 1.0 \\ \hline
        \begin{tabular}[c]{@{}l@{}}$\exp( -0.5842669 + \exp( \cos( 0.7474208 \cdot ( \cos( x_{2} )$ \\ \quad $/ \cos( x_{3} / \exp( \cos( 0.7474208 \cdot x_{2} ) ) ) ) ) ) )$\end{tabular} & 18 & 0.4927 & -0.0303 & 0.9998 & -0.0547 & 0.9999 \\ \hline
        \hline
    \end{tabular}

    } 
\end{table}}

In a first set of experiments, we select \gls{lr} as the victim algorithm $v$, and \gls{qr} as the reference $r$. The expectation is that our approach will discover functions which \gls{qr} can easily fit, and \gls{lr} will not be able to. We then reverse the experiment, using \gls{qr} as the victim and \gls{lr} as the reference. The reverse experiment acts as a negative control, and we expect that no solution will be found. 

The final Pareto fronts obtained for the first configuration are summarized in Table~\ref{tab:pareto-lr-qr}. The $R^2$ test values reported are the best fold for the victim, the worst fold for the reference, and then two mean $\overline{R^2}_{LR}$ and $\overline{R^2}_{QR}$ obtained over the 5-fold cross-validation, to be used as a reference. It is noticeable how, after $f(\mathbf{x})=\cos(x_1)$ at complexity 2, the equations increase in complexity with minimal gains in performance, a possible sign of overfitting. This is further confirmed by the mean $\overline{R^2}$ values, which are nearly identical for complexities of 17 and higher, and only minimally different from the results for complexity 2. 

\begin{figure}[!bt]
    \centering
    \includegraphics[width=0.99\textwidth,trim=3mm 3mm 3mm 8mm,clip]{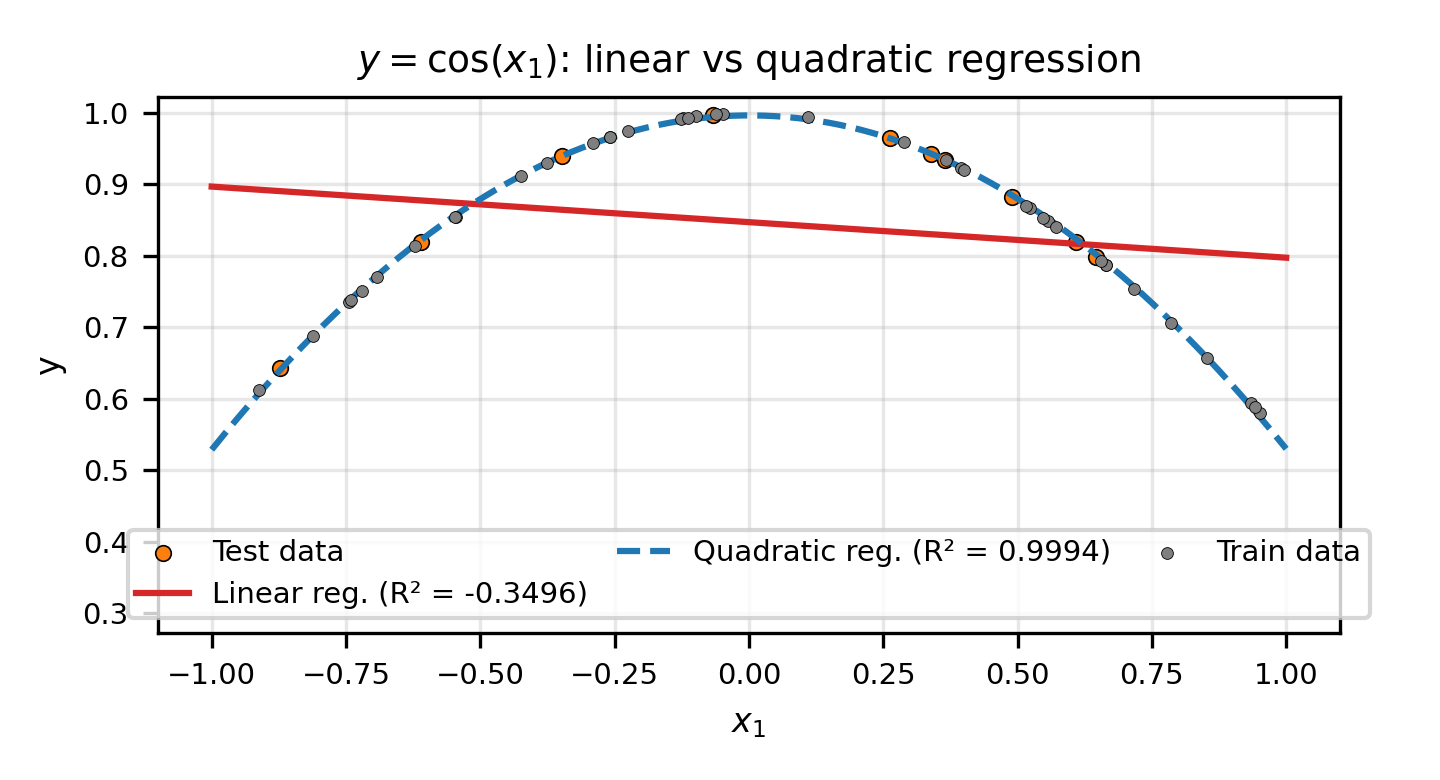}
    \caption{Performance of the models leart by \gls{lr} (red solid) and \gls{qr} (blue dashed) on a dataset obtained randomly sampling $\cos(x_1)$ for $x_1 \in [-1,1]$. In this interval, the function can be approximated quite effectively with a quadratic function, and \gls{qr} learns $\hat{f}(x_1) = 0.9967 + 3.358 \cdot 10^{-4}\, x_1 - 0.4671\, x_1^2$, close to Taylor's approximation $\cos(x) \approx 1 - 0.5x^2$.}
    \label{fig:lr-vs-qr}
\end{figure}

Interestingly, the equation of complexity 17 simplifies to $f(\mathbf{x}) = 0.4652 \cdot x_2^2$, a second-order polynomial, as expected. The equation of complexity 18 is more difficult to decode, but in $\Omega$ it can be approximated by $f(\mathbf{x})=4.48 + 2.40\,x_2^2 - 0.36\,x_3^2~(\overline{R2}=0.9999)$ , another second-order polynomial. Even $f(\mathbf{x})=\cos(x_1)$, since $x_1 \in [-1,1]$ for the domain $\Omega$ used to draw samples, can be closely fit with a second-order polynomial, as $\cos(x_1)\approx1-0.5x_1^2$, see Figure~\ref{fig:lr-vs-qr}. This confirms that the evolutionary search is able to identify functions that \gls{qr} can approximate substantially better than \gls{lr} over the sampled domain.

In contrast, using \gls{qr} as the victim and \gls{lr} as reference does not produce a useful adversarial separation. The resulting Pareto front only contains the trivial linear solution $f(\mathbf{x})=x_2$, which is learned perfectly by both \gls{lr} and \gls{qr}. The marked asymmetry between the two experiments therefore provides the expected positive and negative controls for the proposed framework.

Training for both \gls{lr} and \gls{qr} is quite fast, but performing a cross-validation for each fitness evaluation still takes some time; even with multiprocessing, each run lasted about 25 minutes of wall-clock time on the hardware used for the evaluation.


\subsection{\glsentryshort{xgboost} vs \glsentryshort{svm}}
\label{subsec:xgboost-vs-svm}
\gls{xgboost}~\cite{chen2016xgboost} and \gls{svm}~\cite{cortes1995supportvector} are both widely considered among the state of the art in regression. In this second experiment, we set \gls{xgboost} as the victim and \gls{svm} as the reference. \gls{svm} requires normalization for optimal performance, so features $\mathbf{x}$ and target $y$ were all normalized to zero mean and unit variance, with a transformation learned on the training set of each fold and just applied to the test set. No normalization was applied before running \gls{xgboost}, as the algorithm does \textit{not} need any, being based on an ensemble of \glspl{dt} whose splits are insensitive to absolute feature values. Both algorithms have a sophisticated training process which is considerably more time consuming than the simple \gls{lr}, so the fitness function becomes an obvious bottleneck, with a single run lasting 61 hours of wall-clock time on the same hardware. After the evolutionary runs, the final Pareto fronts are aggregated and reported in Table~\ref{tab:pareto-xgboost-svm}.

{\renewcommand{\arraystretch}{1.4}\setlength{\tabcolsep}{6pt}\begin{table}[!htb]
    \centering
    \caption{Candidate equations from the Pareto front obtained with \gls{xgboost} as the victim and \gls{svm} as the reference.}
    \label{tab:pareto-xgboost-svm}
    \resizebox{\textwidth}{!}{%
        \begin{tabular}{l|c|c|c|c|c|c}
        Candidate function $f(\mathbf{x})$ & Complexity &  $F$ & $R^2_{XGB}$ (best) & $R^2_{SVM}$ (worst) & $\overline{R^2}_{XGB}$ &  $\overline{R^2}_{SVM}$ \\
        \hline \hline
        $x_{3}$ & 1 & 1.067 & 0.9966 & 0.9243 & 0.9883 & 0.9652 \\ \hline
        $\sin( x_{1} )$ & 2 & 1.0324 & 0.9985 & 0.9648 & 0.9948 & 0.9846 \\ \hline
        $3.0015 \cdot 10^{-6} \cdot x_{3} $ & 3 & 0.5666 & -0.0151 & 0.9244 & -0.0845 & 0.9652 \\ \hline
        $-1.3508949 \cdot 10^{-4} \cdot \cos(x_3)$ & 4 & 0.5387 & -0.0095 & 0.9572 & -0.0170 & 0.9652 \\ \hline
        \hline
    \end{tabular}%
    }
\end{table}}



\begin{figure}[!tb]
    \centering
    \includegraphics[width=0.8\textwidth,trim=0mm 0mm 0mm 10mm,clip]{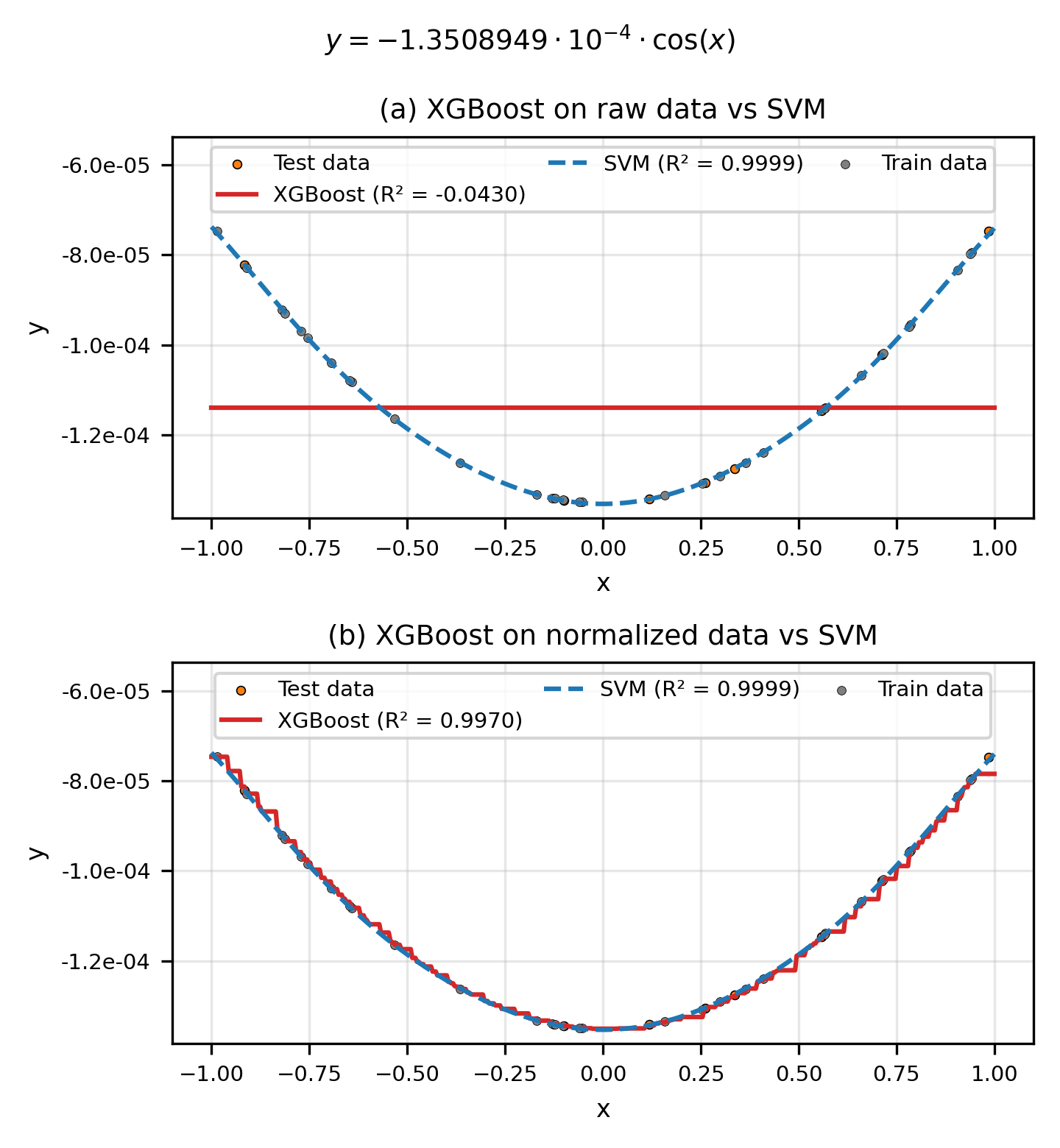}
    \caption{Undesired behavior for \gls{xgboost} uncovered by the proposed approach demonstrated on $y=-1.3508949\cdot 10^{-4}\cdot \cos(x)$. (\textbf{top}) For a function $f$ with small absolute differences in value for the target $y$, \gls{xgboost} just predicts a constant value, while \gls{svm} correctly learns a predictive model of $f$ from the training data. (\textbf{bottom}) The undesired behavior disappears once $y$ is normalized.}
    \label{fig:xgboost-vs-svm}
\end{figure}

Surprisingly, the equations are not very complex, and employ just one of the 3 features; and the most performing ones all include small constants in the range $[10^{-6},10^{-4}]$. Analyzing the performance of \gls{xgboost} on the functions, we can observe a degenerate behavior where the algorithm just predicts a constant value, see Figure~\ref{fig:xgboost-vs-svm}. Investigating the source code of the library, we discover an internal hard-coded numerical constant, $\texttt{kRtEps}=10^{-6}$, which acts as a minimum numerical tolerance in tree-split decisions independently of the user-defined \texttt{gamma} (\texttt{min\_split\_loss}) parameter. In the \texttt{XGBoost 3.4.1} library, this constant is defined in \href{https://github.com/dmlc/xgboost/blob/v3.4.1/include/xgboost/base.h}{\texttt{include/xgboost/base.h}},
where it is described as a ``small eps gap for minimum split decision''. Its documented purpose seems to provide a small numerical margin in the decision of whether a candidate split represents a sufficiently positive improvement.

During tree construction, XGBoost evaluates the improvement produced by candidate splits through the internal quantity \texttt{loss\_chg}. The tree-building procedure implemented in \href{https://github.com/dmlc/xgboost/blob/v3.4.1/src/tree/driver.h}{\texttt{src/tree/driver.h}}
rejects a candidate whenever $\texttt{loss\_chg}\leq \texttt{kRtEps}$, or equivalently, $\texttt{loss\_chg}\leq 10^{-6}$. Thus, even when $\texttt{gamma}=0$, a candidate split must still produce an improvement greater than $10^{-6}$ to pass this internal numerical condition. In practice, this creates an issue where the algorithm's implementation is unable to detect differences between small values in the target label $y$, unless $y$ is normalized. These steps of the algorithm actually explain the behavior shown in Figure~\ref{fig:xgboost-vs-svm}: \gls{xgboost} starts by predicting the mean value $\overline{y}$ of the target in the training set, and then builds the first tree to compensate the errors of this prediction; but since no differences in the values of $y$ are detected, all trees are rejected, and the resulting model is just $\hat{f}_{xgb}(\mathbf{x})=\overline{y}$. Normalizing the target, as previously performed for \gls{svm}, removes the issue.

The common understanding of \gls{xgboost}, however, is that \textit{no} normalization of the features or the target is needed when the base predictors are \glspl{dt}. To the best of our knowledge, \texttt{XGBoost3.4.1}'s documentation\footnote{\texttt{XGBoost3.4.1} docs, \url{https://xgboost.readthedocs.io/en/stable/}} does not mention any necessity of normalizing the target; prompting several commercial \glspl{llm} (accessed on November 6, 2026) with the question ``Do I need to normalize features or target before using \gls{xgboost} for regression?'' seems to confirm this:

\begin{itemize}
    \item \textbf{Claude}: ``No, you generally don’t need to normalize or standardize features for XGBoost. [...] The target variable: \textbf{Scaling the target isn’t required either}, but transforming it can help when it’s heavily skewed, like prices or counts. For example, you could train on \texttt{np.log1p(y)} and invert predictions with \texttt{np.expm1}, or use an objective like \texttt{reg:tweedie} or \texttt{reg:gamma}.''
    \item \textbf{ChatGPT}: ``No, you generally do not need to normalize or standardize features when using XGBoost for regression. [...] \textbf{You generally don't need to normalize the target variable ($y$) either}[...]''
    \item \textbf{Gemini}: ``For standard tree-based XGBoost regression models (booster='gbtree'), \textbf{you do not need to normalize or scale your features or your target variable}.''
\end{itemize}

The only mention of a similar issue we were able to locate is inside an archived discussion on \texttt{XGBoost}'s GitHub repository\footnote{\texttt{XGBoost} issue \#357, Is Normalization Necessary?, \url{https://github.com/dmlc/xgboost/issues/357}}, where a user noticed the behavior; but the conversation has been locked in 2018, with no mention of a resolution, and it seems like the official answer remains that no normalization is required. It is interesting to note that this effect does \textit{not} appear in other implementations of tree-based ensembles like Random Forest~\cite{breiman2001random} or the single \gls{dt} from the Python library \texttt{scikit-learn}~\cite{pedregosa2011scikitlearn}; and not even in different implementations of boosted ensembles, like \texttt{lightgbm}~\cite{ke2017lightgbm}. All other \gls{ml} algorithms based on \glspl{dt} do \textit{not} require the normalization of the target $y$. 


In other words, what the proposed approach uncovered, while potentially already noticed by some users, \emph{is not common knowledge and it is not explicitly described anywhere in the documentation}. This is an undesirable behavior, made more adverse by the fact that all most used implementations of \gls{xgboost} (for \texttt{Julia}, \texttt{Python} and other languages) are wrappers or bindings of this very same \texttt{C++} library (29k stars, 9k forks on GitHub at the time of writing). Normalizing the target $y$ using the same procedure previously adopted for \gls{svm} removes the issue, and raises \gls{xgboost}'s worst fold performance to $\underset{k}{\min }R^2(k)=0.9932$. The bug has been reported to the developers of \texttt{XGBoost} in \href{https://github.com/dmlc/xgboost/issues/12866}{Issue \#12866} and the issue is being corrected in \href{https://github.com/dmlc/xgboost/pull/12876}{Pull Request \#12876}.

\section{Limitations}
\label{sec:limitations}
The proposed approach proved its effectiveness first on a simple toy problem, and then by detecting a weakness in the state-of-the-art regressor \gls{xgboost}, even though the analysis revealed that the source of the issue lies more in the library implementation than in the algorithm proper. Still, despite promising results, it is important to state the limitations of the methodology.

First, hyperparameter settings can considerably change the behavior of a regression algorithm. For example, in the case of \gls{xgboost}, the number of base predictors (\glspl{dt} in this case) can greatly change capacity: \gls{xgboost} with 5 predictors would not be able to fit a target function as well as the same algorithm with 100 predictors. Considering another algorithm, \gls{svm} can show considerable differences in performance depending on the values of $\epsilon$, $C$, and the type of kernel, with \gls{rbf} being only one among several possible choices. So, even if the proposed approach finds a function $f^*$ which is difficult for a victim algorithm with a certain configuration of hyperparameters, the same function might in principle be easily solvable by the same algorithm with different hyperparameters. It is of course possible to perform hyperparameter optimization \emph{inside} the fitness evaluation loop; but the computational cost of a single evaluation, which is already considerable, could become prohibitive; not to mention that even performing hyperparameter optimization requires choices by the user, for example the values to explore in a grid search or the settings of Bayesian optimization for libraries such as Optuna~\cite{akiba2019optuna}.

Another important caveat is that the state-of-the-art regression algorithms are, perhaps obviously, extremely effective. Preliminary experiments performed after circumventing the bug in \gls{xgboost} by normalizing the target $y$, using the same settings as in Section~\ref{subsec:xgboost-vs-svm}, found only minimal differences in performance for equations up to complexity 30, defined as the number of nodes in the \gls{gp} tree. Increasing maximum complexity to 100 led to solutions with better fitness, which are however also extremely difficult to analyze. Thus, there seems to be a trade-off between efficiency and human interpretability: In other words, the proposed approach might be able to find functions that actually are difficult to learn for a victim algorithm, but if they cannot be studied by humans, this would ultimately be impractical.

Some of the methodological choices of the proposed approach could also be reconsidered. For example, what emerges from the results in Tables~\ref{tab:pareto-lr-qr} and \ref{tab:pareto-xgboost-svm} is that using the best performance of $v$ and the worst performance of $r$ in the computation of $F$ (see Equation~\ref{eq:fitness-function}) might not be necessary, and even be slightly deceiving. While using the mean $\overline{R^2}$ could create the issues previously discussed, the median $\widetilde{R^2}$ might be more informative while still resilient to outlier values. Also, limiting $\Omega \in [-1,1]$ is reductive for some functions (e.g. transcendental ones), so enlarging the boundaries to, for example, $\Omega \in [-2\pi,2\pi]$, could make the approach more efficient. 


\section{Conclusions and future work}
\label{sec:conclusions}
Studying the behavior and uncovering the weaknesses of \gls{ml} algorithms is not trivial, due to the opacity and complexity of the trained predictive models. While classic adversarial \gls{ml} frameworks attempt to tackle the issue by finding single samples where a victim algorithm performs incorrect predictions, in this work we propose a novel approach for the evolutionary generation of whole adversarial regression datasets. After a first validation on a simple case study, the proposed approach shows its applicability to real-world configurations by uncovering an undesired behavior in the most used implementation of the state-of-the-art regressor \gls{xgboost}.


Future work will explore further avenues for improving the evolutionary generation of datasets. Since the proposed methodology is independent of the algorithm used to solve the SR problem, alternative approaches, including generative AI methods, could also be investigated. In the current experimental setup, the number of features $d$ and the number of samples $S$ for the datasets obtained sampling the evolved functions $f$ are set to fixed values ($d=3, S=200$). In principle, these two values could also be evolved, adding terms to the fitness function to push for parsimony, e.g. minimizing both number of features and number of samples. The sampling of the functions itself and the split into $k$ folds, which are obtained with simple pseudo-random number generation, could also be evolved. These extensions, however, are not straightforward to implement with out-of-the-box \gls{sr} libraries, and would require an ad-hoc genotype specification. The proposed approach could also in principle be extended to relational data, such as images or text, but it would be necessary to design a representation of the data-generating distribution as a model whose structure or parameters can be both evolved by an \gls{ea} and interpreted by a human, which is far from straightforward; furthermore, deep \glspl{nn} models, which are currently the state of the art for relational data, are also computationally expensive to train, which is impractical for search-and-score approaches. 


\ifarxivstyle
\else
    \begin{credits} 
    \subsubsection{\ackname} One of the authors regrets the choice of American English for the manuscript and suffered tremendously during paper writing. We acknowledge their stoicism.

    \subsubsection{\discintname}
    The authors have no competing interests to declare that are
    relevant to the content of this article.
    
    \alberto{Anna, let me know if you agree with this, and otherwise feel free to modify as you like. A.B. is Omar, C.D. is you, and E.F. is me}
    
    \subsubsection{Statement of contributions}
    AVK and AT designed the method. OLR and AT implemented the code and ran the experiments. OLR, AVK, and AT analyzed the results and wrote the paper together. AT supervised the work. All authors approved the final version and take full responsibility for its content.
    
    
    \subsubsection{Use of AI tools}
    Claude Opus 5.5, Anthropic (accessed in July-November 2026) was used for the initial literature search, to generate part of the \texttt{Julia} code for the experiments, for a first draft of Figure~\ref{fig:overview}, and for all \texttt{Python} code to plot figures and generate \LaTeX~tables starting from experimental data. The authors checked and edited all AI-generated content. No AI tool was used to write the text of the manuscript, including the literature review.
    \end{credits}
\fi

\bibliographystyle{splncs04}
\bibliography{bibliography.bib}

\end{document}